\documentclass[10pt,twocolumn,letterpaper]{article}

\usepackage[pagenumbers]{cvpr} 
\usepackage{subcaption}
\usepackage{caption}
\usepackage{subcaption}
\usepackage{booktabs}
\usepackage{xcolor}
\newcommand{\highlight}[1]{\colorbox{blue!10}{#1}}
\usepackage{float}
\usepackage{amsmath}
\usepackage{amssymb}
\usepackage{algorithm}
\usepackage{algpseudocode} %

\usepackage[utf8]{inputenc} %
\usepackage[T1]{fontenc}    %
\usepackage[dvipsnames]{xcolor}
\usepackage{tabularx}
\usepackage{graphicx}
\usepackage[pagebackref,breaklinks,colorlinks,allcolors=cvprblue]{hyperref}
\hypersetup{colorlinks,
            linkcolor=NavyBlue,
            urlcolor=NavyBlue,
            citecolor=NavyBlue}

\usepackage{pifont}
\newcommand{\cmark}{\ding{51}}%
\newcommand{\xmark}{\ding{55}}%
\usepackage{todonotes}
\usepackage{wrapfig}
\usepackage{multirow}
\usepackage{placeins}
\usepackage{fontawesome5}
\usepackage{comment}
\usepackage{url}            %
\usepackage{booktabs}       %
\usepackage{amsfonts}       %
\usepackage{nicefrac}       %
\usepackage{microtype}      %
\usepackage{xcolor}         %
\usepackage{xspace}
\usepackage{amsmath}
\usepackage{graphicx} 

\let\originalleft\left
\let\originalright\right
\renewcommand{\left}{\mathopen{}\mathclose\bgroup\originalleft}
\renewcommand{\right}{\aftergroup\egroup\originalright}
\newcommand{\Fig}[2][]{\hyperref[{#2}]{Figure~\ref*{#2}#1}}
\newcommand{\fig}[2][]{\hyperref[{#2}]{Fig.~\ref*{#2}#1}}
\newcommand{\figs}[1]{\hyperref[{#1}]{Figs.~\ref*{#1}}}  %
\newcommand{\Tab}[1]{\hyperref[{#1}]{Table~\ref*{#1}}}
\newcommand{\tab}[1]{\hyperref[{#1}]{Table~\ref*{#1}}}
\newcommand{\Eqn}[1]{\hyperref[{#1}]{Equation~\ref*{#1}}}
\newcommand{\eqn}[1]{\hyperref[{#1}]{Eq.~\ref*{#1}}} %
\renewcommand{\sec}[1]{\hyperref[{#1}]{Sec.~\ref*{#1}}} %
\newcommand{\Sec}[1]{\hyperref[{#1}]{Section~\ref*{#1}}} %
\newcommand{\supp}[1]{\hyperref[{#1}]{Suppl.~\ref*{#1}}}
\newcommand{\app}[1]{\hyperref[{#1}]{App.~\ref*{#1}}}
\newcommand{\App}[1]{\hyperref[{#1}]{Appendix~\ref*{#1}}}

\newcommand{\Alg}[1]{\hyperref[{#1}]{Algorithm~\ref*{#1}}}

\usepackage{xspace}
\makeatletter
\DeclareRobustCommand\onedot{\futurelet\@let@token\@onedot}
\def\@onedot{\ifx\@let@token.\else.\null\fi\xspace}

\makeatother

\definecolor{cvprblue}{rgb}{0.21,0.49,0.74}
\definecolor{stanfordred}{RGB}{140, 21, 21}

\def\paperID{} %
\def\confName{CVPR}
\def\confYear{2025}
\newcommand{\method}{\mbox{\textsc{CTRL-O}}\xspace}  %
\newcommand{\DINOSAUR}{\textsc{Dinosaur}\xspace}

\title{\method: Language-Controllable Object-Centric Visual Representation Learning}

\author{%
  \textbf{Aniket Didolkar}$^{1,2\ast}$ \quad \textbf{Andrii Zadaianchuk}$^{3\ast}$\textsuperscript{\dag} \quad  \textbf{Rabiul Awal}$^{1,2}$\thanks{denotes equal contribution, order is determined by flipping a coin} \\  \textbf{Maximilian Seitzer}$^{4}$ \quad \textbf{Efstratios Gavves}$^{3, 5}$ \quad \quad \textbf{Aishwarya Agrawal}$^{1,2}$\thanks{denotes equal advising contributions}   \\
  $^{1}$Mila - Quebec AI Institute, $^{2}$Université de Montréal \\ $^{3}$University of Amsterdam, The Netherlands $^{4}$University of Tübingen \\ $^{5}$Archimedes/Athena RC, Greece\\ 
    \raisebox{-0.35\height}{\includegraphics[width=1.25em,height=1.25em]{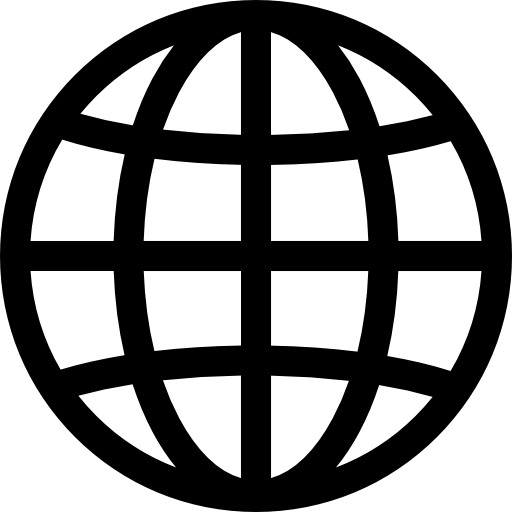}} \hspace{0.2em}
    \large{\url{https://ctrl-o-paper.github.io}}
}

\begin{document}
\maketitle
\begin{abstract}
Object-centric representation learning aims to decompose visual scenes into fixed-size vectors called ``slots'' or ``object files'', where each slot captures a distinct object. Current state-of-the-art object-centric models have shown remarkable success in object discovery in diverse domains, including complex real-world scenes. However, these models suffer from a key limitation: they lack controllability. Specifically, current object-centric models learn representations based on their preconceived understanding of objects,
without allowing user input to guide which objects are represented. Introducing controllability into object-centric models could unlock a range of useful capabilities, such as the ability to extract instance-specific representations from a scene. In this work, we propose a novel approach for user-directed control over slot representations by conditioning slots on language descriptions. The proposed \textsc{\textbf{C}on\textbf{TR}o\textbf{L}lable \textbf{O}bject-centric representation learning} approach, which we term \method, achieves targeted object-language binding in complex real-world scenes without requiring mask supervision. Next, we apply these controllable slot representations on two downstream vision language tasks: text-to-image generation and visual question answering. 
The proposed approach enables instance-specific text-to-image generation and also achieves strong performance on visual question answering.
\end{abstract}

\section{Introduction}
\label{sec:intro}

Object-centric representation learning aims to decompose a visual scene into its constituent entities or objects and represent each entity as a distinct vector called a \textit{slot}.
Slot-based representations are inherently compositional and support many complex downstream tasks such as dynamics modeling~\citep{alias2021neural, wu2023slotformer, manasyan2025temporally}, control~\cite{zadaianchuk2020smorl,driess2023palme,didolkar2024cycle,haramati2024entitycentric}, and reasoning~\cite{assouel2022objectcentric,mamaghan2024exploringVQA}. Moreover, studies in cognitive neuroscience \cite{Spelke2000CoreKnowledge, Pinker1984VisualCognition}  have shown that human perception uses mechanisms akin to slot-based representations. \looseness=-1

\begin{figure}
    \small
    \centering
    \includegraphics[width=\linewidth]{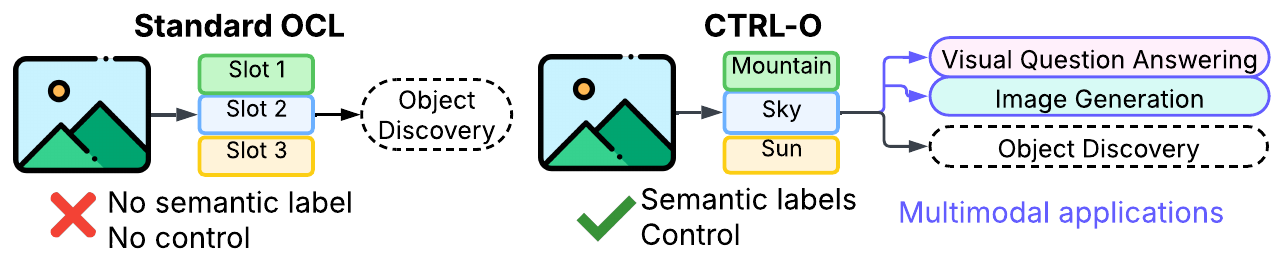}
    \caption{Left: Standard object-centric learning~(OCL) assigns arbitrary slots with no control. Right: \method introduces language-based control, enabling specific object targeting and multimodal applications.\looseness=-1}
    \label{fig:teaser}
\end{figure}

Existing unsupervised object-centric models~\citep{eslami2016attend, Engelcke2020GENESIS, Locatello2020SlotAttention, seitzer2023bridging, manasyan2025temporally, didolkar2024cycle} can successfully decompose complex real-world visual scenes. However, they face a fundamental limitation: they lack control over the object representations. While these models expose control over the \textit{number} of scene parts, they do not allow users to extract specific object representations within a scene (e.g., specified by user queries in the form of language or position markers). For instance, a user can specify that a scene should be decomposed into $K$ slots, but they cannot direct a given slot to bind to a particular object of interest such as ``a cat'' or ``a black purse''. 

This lack of control over the semantic content of the representation can be limiting, as this restricts the model always to extract a fixed decomposition of a scene based on its own preconceived understanding of objects and parts. 
Such rigidity can be problematic for applications that require representations at varying granularity, such as extracting the representation of a car wheel instead of the entire car, or vice versa.
\looseness=-1

Moreover, many downstream tasks may require or benefit from the knowledge of the semantic content in the slots. Due to the unsupervised nature of existing models, there is no way to identify the content of a slot without manually checking its corresponding mask. For example, if the user needs representations of the cat and a dog in a particular image to answer a question, they would have to manually inspect masks of all discovered objects to pick the corresponding slots.\looseness=-1

To address these limitations, we propose to inject controllability into object-centric representation learning. We achieve this by querying the model to represent specific objects in the image. Specifically, these queries condition the slot vectors to guide them to the objects described by the query. The queries can be in the form of natural language (such as object category names or referring expressions). The main challenge is to ensure that the slots conditioned on a specific query bind to the object referred to by that query. We term the challenge of binding slots to specific objects the \textit{visual grounding problem}~\cite{kazemzadeh2014referitgame,xu2015show,greff2020binding, perez2018film}. We find that this problem is not trivial and introduce a novel controllable object-centric model --- \method~--- to solve it. 
In our experiments, we demonstrate that the proposed approach can successfully bind slots to objects specified by user queries containing object categories or referring expressions in complex real-world scenes with limited supervision. In addition, we demonstrate the usefulness of the extracted controllable representations for two downstream tasks: visual question answering and instance-controllable image generation. Our contributions are as follows: 
\begin{itemize}
    \item We introduce CTRL-O, a novel method to learn controllable object-centric representations via user-defined inputs. %
    \item We demonstrate that this approach supports extracting representations for complex reference expressions, enabling precise part specification within the representation.
    \item We validate the effectiveness of \method on two real-world downstream tasks: Instance-Controllable Image Generation and Visual Question Answering.\looseness=-1   %
\end{itemize}
\section{Related Works}

\paragraph{Object-Centric Representation Learning} Unsupervised object-centric representation learning (OCL) has gained a lot of interest in recent years~\citep{eslami2016attend,greff2019multi,Engelcke2020GENESIS,Locatello2020SlotAttention,  biza2023invariant,  seitzer2023bridging, didolkar2024cycle, kipf2022conditional, zadaianchuk2023videosaur, manasyan2025temporally,  didolkar2024zero}.  OCL aims to extract individual representations for various entities in unstructured sensory inputs such as images. Slot Attention~\citep{Locatello2020SlotAttention} introduces an attention-based mechanism to decompose images into object-centric representations. \DINOSAUR~\citep{seitzer2023bridging} builds upon this by utilizing self-supervised DINO features~\citep{Caron2021DINO, oquab2023dinov2} to enhance unsupervised object discovery. While \DINOSAUR can effectively identify objects in real-world data~\citep{Lin2014COCO}, it lacks mechanisms for top-down control over the representations.
In contrast, \method provides controllable OCL by incorporating language-based control queries, allowing for flexible guidance with minimal supervision during training.
Some works~\citep{kipf2022conditional, kori2023unsupervised} have explored conditioning mechanisms in object-centric models.  SAVi~\citep{kipf2022conditional} uses bounding boxes for the initial frame of a video for conditioning. CoSA~\citep{kori2023unsupervised} conditions on learned vector representations. These methods are often limited to specific forms of conditioning and are primarily evaluated on synthetic datasets, while our method can handle many forms of conditioning on real-world data. Finally, several recent works connect object-centric representations with language~\citep{wang2020language, fan2023unsupervised, kim2023shatter}. These works connect object representations with language post-hoc, assigning language labels to discovered slots. In contrast, \method integrates language and point conditioning directly into the learning process, allowing a user to control what representations should be extracted.
\looseness=-1

\paragraph{Downstream tasks with object-centric representations} There has been limited work exploring the applicability of object-centric models to downstream tasks. Slotformer~\citep{wu2023slotformer} uses the learned slots for world modelling and video question answering. \citet{zadaianchuk2020smorl}, \citet{yoon2023investigation} and \citet{didolkar2024cycle} investigate the applicability of object-centric representations for learning RL policies in simple environments such as Atari~\citep{mnih2013playing}. One drawback of these works is that they mainly consider synthetic environments and toy tasks; thus, their applicability is limited. In contrast, in this paper, we consider downstream applications in complex real-world environments. There are only a few works that study applications of object-centric representations in real-world settings. Mamaghan et al \cite{mamaghan2024exploringVQA} investigate the application of object-centric representations in Visual Question Answering. We consider them as a baseline for our experiments on VQA. Slot Diffusion \cite{wu2023slotdiffusion} and Stable LSD \cite{jiang2023object} use object-centric representations for generating real-world images. However, both these approaches lack controllability; hence, it is difficult to specify any conditioning information or control the images that these approaches generate. In contrast, we demonstrate in Section \ref{sec:imagegen}, that \method, when used for image generation, provides fine-grained control over the image generation.\looseness=-1

\section{Method} \label{sec:method}
\begin{figure*}[t]
\centering
\includegraphics[width=\textwidth]{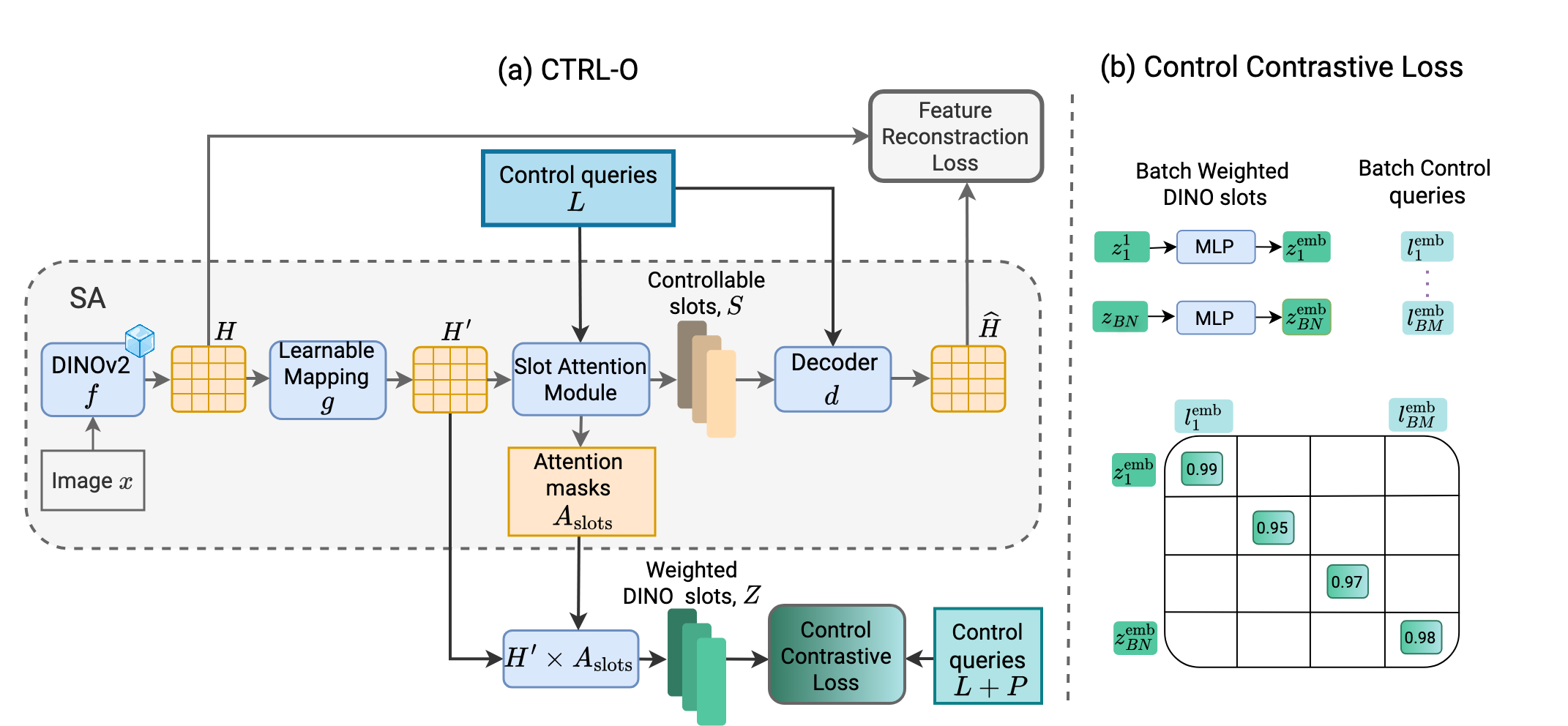} \\
\caption{(a) Overview of \method architecture. An input image is processed by a frozen DINOv2 ViT model $f$, yielding patch features $H$. These features are then transformed into $H'$ by a learnable transformer encoder $g$ to align the feature space with the control queries. The control queries are introduced in the Slot Attention (SA) module, which guides the grouping of the encoded features into slots $S$. The initial slots in the SA module are conditioned with the control queries. Finally, an MLP decoder $d$, conditioned on control queries, reconstructs the DINOv2 features. (b) To ensure that slots utilize query information to represent specific objects, we apply a contrastive loss between control queries and the Slot Attention-modulated weighted DINO features $A_{\text{slot}}$ (referred to as weighted DINO slots).}
\label{fig:framework}
\end{figure*}
In this section, we describe the proposed approach for injecting controllability into existing object-centric models. %
We present a visual depiction of our method in \fig{fig:framework}. 

In our setup, the input consists of an image $X$ and user-defined queries embedded into vectors $L = \{l_j \in \mathbb{R}^{D_{\text{emb}}}\}_{j=1}^M$. The expected object-centric representation of the image $X$ is a set of slots $S = \{s_i\in \mathbb{R}^{D_{\text{slot}}}\}_{i=1}^N$. 
The first $M$ slots (we assume that $M\leq N$) should represent the object identified by the corresponding queries, while the remaining slots represent the unspecified parts of the scene. This way, the obtained representation is a complete decomposition of the whole image $X$, while still containing the parts corresponding to the user-specified queries $L$.

We consider controllability in the form of \textit{language queries}. We rely on the user to provide free-form text specifying object categories or object referring expressions whose visual representations are sought after. We encode this text into a fixed-sized vector embedding using LLM2Vec~\cite{behnamghader2024llm2vec} with LLaMA-3-8BLLaMA-3-8B~\cite{dubey2024llama3herdmodels}  to obtain these embeddings. These language embeddings comprise the queries we feed into the model.\looseness=-1

\subsection{\method Architecture}

\textbf{Background} We base the proposed approach on \DINOSAUR \cite{seitzer2023bridging}. \DINOSAUR uses the Slot Attention module~\cite{Locatello2020SlotAttention} for object discovery. Slot Attention is an attention-based differentiable clustering procedure which, given a grid of features $H = \{h_k\}_{k=1}^K = f(X)$ obtained from an encoder $f$ (we use DINOv2~\citep{oquab2023dinov2}) applied to an image $X$, outputs a set of slots  $S$ such that each slot represents a distinct object in the image (see \app{sec:dinosaur} for a detailed \DINOSAUR description).

\textbf{Query-based Slot Initialization} We are solving the visual grounding problem: given the query corresponding to an object in the image, 
we want a slot to bind to exactly that object. 
One straightforward way to enforce grounding is to condition the slots directly on the query corresponding to each object.  Specifically, we achieve this by adding the object query $l_i$ to one of the slots (see \fig{fig:framework},  input to the Slot Attention Module). This approach is similar to SAVi~\cite{kipf2022conditional}, which conditions each slot on the objects' center of mass. In our experiments, we find that simply conditioning the slots on the queries does not lead to correct grounding; hence, a stronger signal is needed to ensure proper grounding.

\textbf{Decoder Conditioning} Similar to \DINOSAUR, we use a broadcast MLP decoder, separately decoding each slot into patch features. We empirically find that conditioning the decoder on the corresponding control queries improves language grounding (concrete evaluation presented in \Tab{tab:grounding_ablation}. To implement this, we concatenate the resulting slots with the control queries and pass them through an MLP whose output is fed into the patch decoder as shown in \fig[ (a)]{fig:framework}.

\subsection{Control Contrastive Loss to Enforce Grounding} 
To enforce grounding, we introduce a contrastive loss, as illustrated in \fig[ (b)]{fig:framework}. The intuition behind this objective is that if a slot $s_i$ is conditioned on a query $l_i$ corresponding to the object $o_i$, then we want the encoder features corresponding to the slot $s_i$ to be close in embedding space to the query $l_i$. To obtain the features corresponding to slot $s_i$, we spatially aggregate the features output by the mapping network (learnable mapping $g$ in \fig[ (a)]{fig:framework}) by weighting them with the attention scores of slot $s_i$, obtained from the last iteration of slot attention: $z_i = \sum_{k=1}^K a_{ik} h_{k}$, where $a_{ik}$ denotes the attention score of slot $s_i$ on feature $h_k$. We further process $z_i$ using an MLP to output $z^{\text{emb}}_i$, which is used in the contrastive loss. 
Note that we do not directly use the slots for the contrastive loss because the loss can be trivially satisfied by the slots as they are conditioned on the control queries, which are the targets for the contrastive loss. 
For the contrastive loss, we use $(z^\text{emb}_i, l_i)$ as positive pairs which should be similar, and $(z^\text{emb}_i$, $l_t)$ as negative pairs which should be dissimilar, with $t \neq i$.
For the negatives, we consider all conditioning queries across the entire batch. Let there be $T$ conditioning queries in the batch. The loss for a single sample is then formalized as:
\begin{equation}
    \mathcal{L}^{l}_{CC} = - \sum_{i=1}^M \log\frac{\exp(z^{\text{emb}}_i \cdot l_i/\tau)}{\sum_{t=1}^T \exp(z^{\text{emb}}_i \cdot l_t/\tau)}
\end{equation}
Here, $\tau$ is the temperature, which is set to $0.1$. We assume two training regimes: when only language queries $l_i$ are available and when both language queries $l_i$, and center-of-mass point queries $p_i$ are provided during training. In the first regime, we define control contrastive loss $\mathcal{L}_{CC}$ as simply $\mathcal{L}^{l}_{CC}$. In the second regime, control contrastive loss $\mathcal{L}_{CC}$ is defined as the sum of two losses with language and point queries: $\mathcal{L}_{CC} = \mathcal{L}^{l}_{CC} + \mathcal{L}^{p}_{CC}$. We incorporate control contrastive loss $\mathcal{L}_{CC}$ in addition to the feature reconstruction loss from \DINOSAUR. 
For additional implementation details, see \app{app:sec:implement_details}.
\looseness=-1

\section{Experiments}

\begin{figure}
    \centering
    \includegraphics[width=\linewidth]{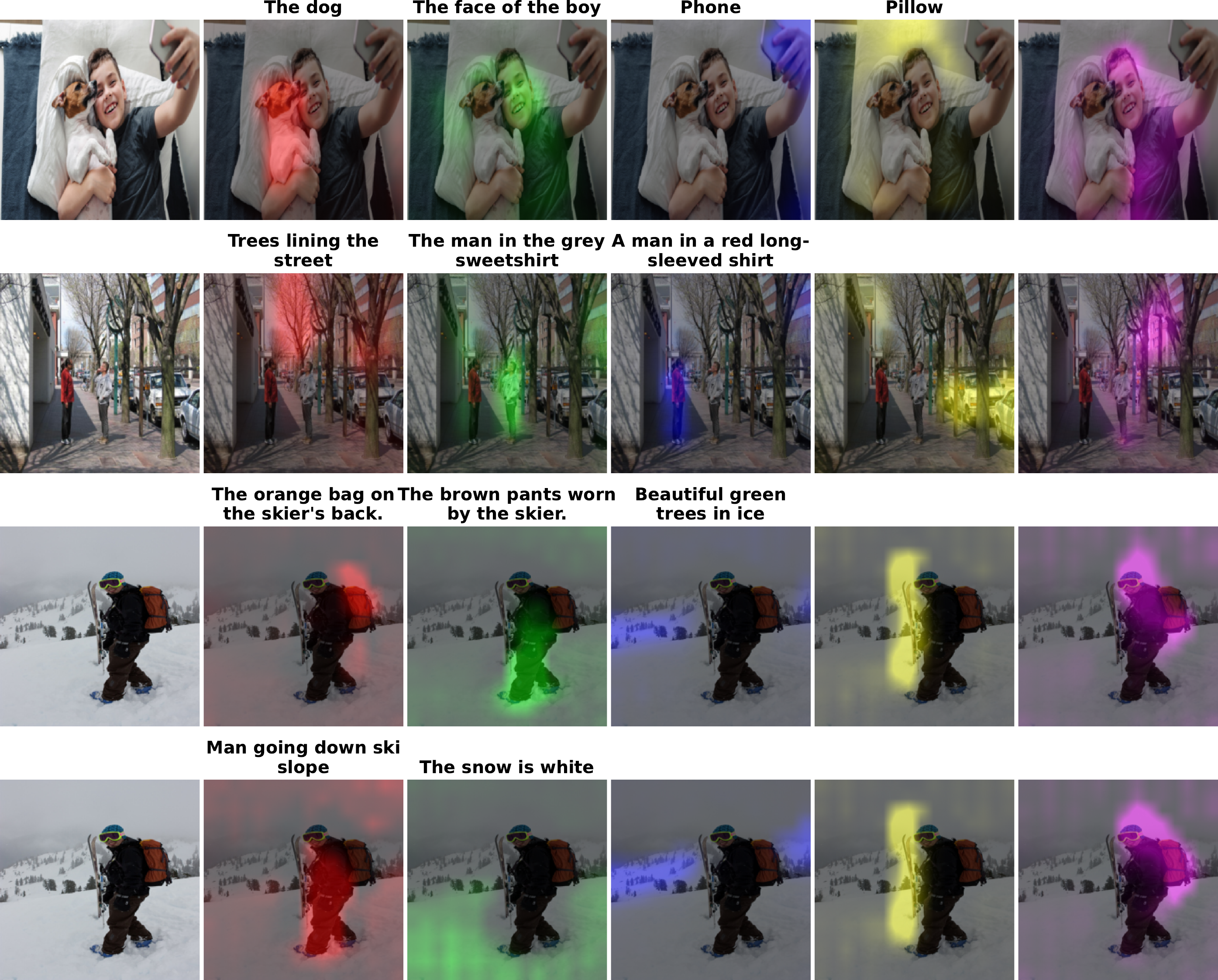}
    \caption{\textbf{Referring Expression Controllability on Visual Genome}. Visualization \method with free-form queries. The original image (left) and predicted segmentation masks are shown, with conditioning phrases presented above the corresponding segmented image; unconditioned slots have no phrase.}
    \label{fig:vg_demo}
\end{figure}

\begin{table}[t]
    \centering
    \small
    \setlength{\tabcolsep}{0.8mm}
    \renewcommand{\arraystretch}{1.5}
    \begin{tabular}{cccc|cccc}
        \toprule
        \bf Slot Init. & \bf GT Masks  & \bf CL & \bf DC & \bf Binding Hits & \bf FG-ARI & \bf mBO \\ 
        \midrule
        \ding{51} & \ding{51} & \ding{55} & \ding{55} & \textbf{71.2} & \textbf{69.8} & \textbf{35.4}   \\
        \hline
        \ding{51} &  \ding{55}  & \ding{55} & \ding{55} &  8.1  &  34.52 & 22.42  \\
        \ding{51} &  \ding{55} & \ding{55} & \ding{51} & 10.11 & 43.83 &  25.76 \\
        \ding{51} &  \ding{55}  & \ding{51} & \ding{55} & 56.3 & 44.8 & \highlight{27.3}   \\
        \ding{51} &  \ding{55} & \ding{51} & \ding{51} & \highlight{61.3} & \highlight{47.5} & 27.2  \\
        \bottomrule
    \end{tabular}
    \caption{\textbf{\method Model Component Ablation for Grounding}. Importance of various components for achieving strong grounding. We use COCO \textit{train} set for training and \textit{val} set for evaluation. CL = Contrastive Loss, DC = Decoder Conditioning.}
    \label{tab:grounding_ablation}
\end{table}

\begin{table}[t]
    \centering
    \small
    \setlength{\tabcolsep}{1.6 pt}
    \renewcommand{\arraystretch}{1.2}
    \begin{tabular}{llcc}
        \toprule
        \multirow{1}{*}{\bf Approach} & \multirow{1}{*}{\bf Model} & \bf FG-ARI & \bf mBO \\
        \midrule
        \multirow{4}{*}{Unsup.} & \DINOSAUR (MLP Dec.)~\citep{seitzer2023bridging} & 40.5 & 27.7 \\
        & \DINOSAUR (TF.~Dec.)~\citep{seitzer2023bridging} & 34.1 & \highlight{31.6} \\
        & Stable-LSD~\citep{jiang2023object} & 35.0 & 30.4 \\
        & SlotDiffusion~\citep{wu2023slotdiffusion} & 37.3 & 31.4 \\
        \hline
        \multirow{2}{*}{Weak Sup.} & Stable-LSD (Bbox Supervision)~\cite{singh2024guidedlatentslotdiffusion} & - & 30.3 \\
        & \method (Trained on COCO) & \highlight{47.5} & 27.2 \\
        \bottomrule
    \end{tabular} 
    \caption{\textbf{Object Discovery Performance.} Comparison of \method with unsupervised and weakly-supervised object-centric approaches on the COCO dataset.}
    \label{tab:segmentation_comparison}
\end{table}

In this section, we first show that \method learns to bind to the right regions in the image given complex natural language queries. Next, we tackle two downstream tasks --- instance-specific image generation and visual question answering --- using a pretrained \method model.\looseness=-1 

\subsection{Grounding Object-Centric Models}\label{sec:gounding}
We study \method across two axes: 1)~\textbf{Object Discovery} --- How well can \method discover and represent each object separately in a scene?, and 2) \textbf{Grounding} --- Can the slot conditioned on some language query $l_j$ bind to the region specified by that language query?

\paragraph{Metrics} To evaluate object discovery, we use standard metrics such as adjusted rand index (ARI) \cite{hubert1985comparing} and mean best overlap (mBO) \cite{Arbelaez2014MCG}. To measure grounding, we introduce a new metric called \textit{Binding Hits} which measures the grounding accuracy of the conditioned slots. Refer to \app{app:sec:metrics} for more details regarding these metrics.

\paragraph{Datasets and Training Details} We use COCO \cite{Lin2014COCO} and Visual Genome (VG)~\cite{krishna2016visualgenomeconnectinglanguage} as our main datasets of study. 
COCO contains category annotations spanning 91 different categories while VG contains region descriptions. COCO contains object annotations along with corresponding segmentation masks; we use it for quantitative evaluation of \method on object discovery and grounding. VG does not contain segmentation masks; hence, we only evaluate on it qualitatively.%
\begin{table*}[t!]
\setlength{\tabcolsep}{0.2pt}
\normalsize	
\centering
\scalebox{0.9}{
\begin{tabular}{p{3.3cm}<{}|p{1.2cm}<{\centering}|p{2.8cm}<{\centering}|p{1cm}<{\centering}|p{1cm}<{\centering}p{1cm}<{\centering}p{1cm}<{\centering}|p{1cm}<{\centering}p{1cm}<{\centering}p{1cm}<{\centering}|p{1cm}<{\centering}}
\toprule
\multirow{2}{*}{Methods} & \multirow{2}{*}{Sup.} & \multirow{1}{*}{Image-text}
& \multirow{1}{*}{Fine}
& \multicolumn{3}{c|}{RefCOCO}
&\multicolumn{3}{c|}{RefCOCO+} 
&\multicolumn{1}{c}{Gref} \\
                                                   &   & pretraining dataset  & tuning & \textit{val} & \textit{testA} & \textit{testB}   & \textit{val} & \textit{testA} & \textit{testB}  & \textit{val}\\ \midrule
\multirow{2}{*}{GroupViT~\cite{xu2022groupvit}}     & 
\multirow{2}{*}{$\mathcal{T}$} & \multirow{2}{*}{CC12M+YFCC}           & \xmark &7.99   & 6.16  & 10.51 & 8.49  & 6.79  & 10.59 & 10.68 \\
                                                   &   &                                      & \cmark & 10.82 & 11.11 & 11.29 & 11.14 & 10.78 & 11.84 & 12.77  \\ \midrule 
\multirow{2}{*}{MaskCLIP~\cite{zhou2022maskclip}}    & \multirow{2}{*}{$\mathcal{T}$} & \multirow{2}{*}{WIT}      & \xmark & 11.52 & 11.85 & 12.06 & 11.87 & 12.01 & 12.57 & 12.74 \\ 
                                                   &   &                                      & \cmark & 19.45 & 18.69 & 21.37 & 19.97 & 18.93 & 21.48 & 21.11  \\ 
                                                   \midrule %

\multirow{1}{*}{Shatter \& Gather~\citep{kim2023shatter}} & \multirow{1}{*}{$\mathcal{T}$} & VG  & \xmark & 21.80 & 19.00 & 24.96 & 22.20 & 19.86 & \highlight{24.85} & 25.89  \\

\midrule

\multirow{1}{*}{\method}  & \multirow{1}{*}{$\mathcal{T}$}  & VG  & \xmark & 21.80 & 20.10 & 21.57 & 21.90 & 21.54 & 21.36 & 25.32  \\

\multirow{1}{*}{\method}  & \multirow{1}{*}{$\mathcal{T+P}$}  & VG  & \xmark & \highlight{28.2} & \highlight{33.13} & \highlight{27.05} & \highlight{25.87} & \highlight{30.58} & 22.58 & \highlight{30.50}  \\
\bottomrule
\end{tabular}
}
\caption{
\textbf{Referring expression segmentation} Comparison with weakly-supervised reference expression segmentation methods (Shatter \& Gather) and open-vocabulary segmentation methods (GroupViT and MaskCLIP). The results on three datasets are reported in mIoU (\%). Fine-tuning \cmark means that the model is trained with the image-text pairs of the target benchmark; otherwise, the model is trained on the image-text pretraining dataset, and applied to the reference datasets zero-shot.
}
\label{tab:comp_open}
\end{table*}
As several images in COCO contain multiple instances of the same object category, conditioning multiple slots on the same category name can be ambiguous for the model. Also, such conditioning poses problems for reliably computing the Binding Hits metric. Therefore, to disambiguate multiple instances of the same object category, we condition the slots on both category names and center of mass coordinates. We embed the language query using Meta-LLaMA-8B and the center of mass coordinates using a 2-layered MLP and concatenate them into a conditioning vector. \looseness=-1 

\method is trained for 300k steps on VG and COCO datasets with a batch size of $128$ and Adam with $0.0004$ learning rate. For VQA downstream task~(\sec{sec:vqa}), we use a batch size of $32$ and AdamW with $5e-4$ learning rate.\looseness=-1

\paragraph{Object Discovery~(\Tab{tab:segmentation_comparison})} We compare \method to various unsupervised and one weakly-supervised object discovery method. All the methods considered in \tab{tab:segmentation_comparison} apply Slot Attention to the features of a pretrained encoder to extract slots. Following \DINOSAUR, this has become the standard in unsupervised object discovery. The weakly-supervised approach, Stable LSD (w/ bbox supervision) \cite{singh2024guidedlatentslotdiffusion}, uses bounding boxes to supervise the Slot Attention alpha masks. \method conditions the slots on language and center of mass queries and also uses the same information for contrastive loss. Therefore, \method can be classified as a weakly-supervised approach. Note that we do not use any of the guidance information to directly supervise the Slot Attention masks; thus, our form of supervision is weaker as compared to Stable LSD (w/ bbox supervision). In \tab{tab:segmentation_comparison}, we show that \method outperforms all unsupervised approaches in terms of ARI but lags behind in terms mBO. The lower performance in terms of mBO can be attributed to the MLP decoder of the underlying \DINOSAUR model, which also obtains a lower mBO. We find that a transformer decoder \cite{seitzer2023bridging} or a diffusion decoder \cite{singh2024guidedlatentslotdiffusion, wu2023slotdiffusion, jiang2023object} results in sharper masks as compared to the MLP decoder. This experiment verifies that \method can discover objects in complex natural scenes. Next, we evaluate whether it can bind to the region specified by the control queries. 
\begin{figure*}[th!]
    \centering
    \includegraphics[trim={0 1.3cm 1cm 0cm},clip, width = 0.9\linewidth]{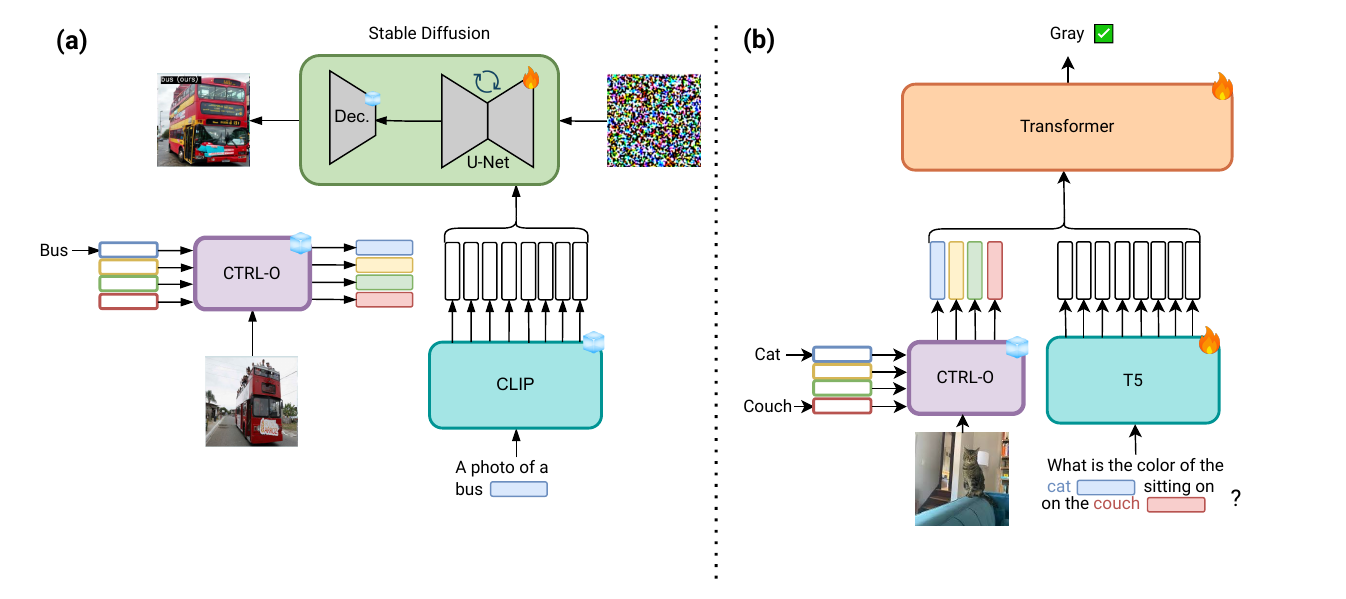}
    \caption{ (a) \textbf{Instance Specific Image Generation}: We query an image to extract instance slot representations, which are then input into a Stable Diffusion model with the caption to generate the image. (b) \textbf{Visual Question Answering}: Slots are extracted from noun chunks or referring expressions in the question, then embedded into the text and input into the language model. \includegraphics[height=1em]{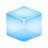} = frozen; \includegraphics[height=1em]{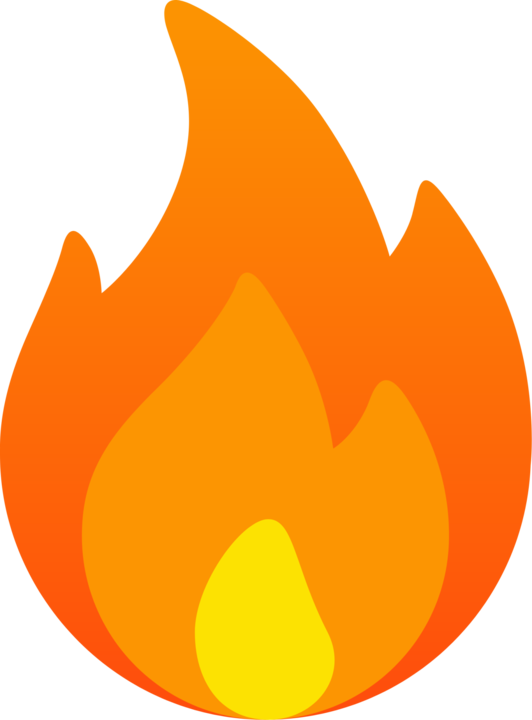} = trainable.} \label{fig:downstream_tasks}
\end{figure*}

\paragraph{Grounding Object Categories~(\Tab{tab:grounding_ablation})}  Controllability is a new paradigm for object-centric models that has not been explored before. Hence, there are no direct baselines with which we can compare. Instead, we try to demonstrate the difficulty of the grounding problem and ablate over the components introduced in \sec{sec:method} to understand their importance in achieving good grounding. We use the COCO dataset for this evaluation. \Tab{tab:grounding_ablation} presents the results for various ablations. 
To obtain an upper bound for grounding performance, we train a (fully supervised) \method model by directly predicting the ground truth masks (first row in \tab{tab:grounding_ablation}).
While such a supervised baseline achieves strong segmentation performance (as indicated by ARI and mBO), it still cannot achieve perfect grounding (close to 100\% Binding Hits), highlighting the difficulty of the grounding problem. 
Out of the components introduced in \sec{sec:method}, the control contrastive loss is the most crucial component for achieving good grounding accuracy, followed by decoder conditioning. Without the contrastive loss, the model has no incentive to utilize the queries; hence, no binding is emerging.\looseness=-1

\paragraph{Grounding Referring Expressions~(\Fig{fig:vg_demo})}
For COCO, we achieved controllability through both center of mass and category information. However, this approach is limited: COCO has a fixed number of categories, which affects generalizability. To overcome this issue, we can rely on \emph{referring expressions}, where a user can refer to the target object using free-form natural language queries. To incorporate this ability, we use the VG dataset \cite{krishna2016visualgenomeconnectinglanguage}. Since VG does not provide segmentation masks, we only evaluate \method qualitatively in this dataset. %

We present the qualitative evaluation on VG in \fig{fig:vg_demo}. We visualize the attention regions of each slot for a random sample of examples. We observe that slots conditioned on phrases bind to the object referred to in the phrase, while the unconditioned slots bind to the other objects not used for conditioning. The last two rows of the visualization demonstrate that \method can decompose the same scene at different levels of granularity, such as, just the leg (corresponding to the query ``The brown pants worn by the skier''), just the bag (``The orange bag on the skier's back''), or the entire skier (``Man going down ski slope'').\looseness=-1 

\paragraph{Referring Expression Segmentation Evaluation~(\Tab{tab:comp_open})} We evaluate \method trained on VG on referring expression segmentation on the RefCOCO, RefCOCO+, and Gref datasets. This is a zero-shot evaluation since these datasets were not used for training. We compare to various referring expression segmentation baselines, using mIoU between the predicted and ground truth mask as the metric. The most relevant baseline is Shatter \& Gather (SaG) \cite{kim2023shatter}, which also employs slot attention to extract slots from the image, followed by cross attention to associate the referring expression with a slot. One limitation of SaG is that the referring expression does not directly influence the slot extraction process. This can be problematic for cases where the referring expression refers to an object not extracted by Slot Attention. This is not the case for \method, as the referring expression directly influences the slot extraction. Moreover, all the baselines mentioned in \tab{tab:comp_open} can only process a single referring expression per forward pass, while \method can process multiple referring expressions in parallel by conditioning multiple slots on the corresponding expressions. We find that \method outperforms all the baselines. However, \method by default uses language queries and center of mass annotations ($\mathcal{T} + \mathcal{P}$) while the baselines use only language queries as weak supervision. To ensure fairness, we also evaluate \method with only language queries ($\mathcal{T}$), which achieves competitive performance with SaG. Implementation details for this variant are in \app{app:ctrl_o_language_only}.\looseness=-1

\subsection{Unlocking New 
OCL Capabilities 
with \method}
In this section, we show that \method can be used for \emph{instance controllable image generation} (\fig[(a)]{fig:downstream_tasks}) --- a use case where existing object-centric methods fail. We also demonstrate how \method can be used in a novel way to improve over existing object-centric methods in Visual Question Answering~(VQA) (\fig[(b)]{fig:downstream_tasks}). In these experiments, we use \method pre-trained on both VG and COCO with language-only conditioning (referring expressions in VG and Object Categories in COCO). %

\subsection{Instance Controllable Image Generation} \label{sec:imagegen}

In this section, our goal is to demonstrate that object-centric representations obtained from \method can be used for controllable image generation. Specifically, we aim for control in the space of instances where specific instances can be extracted from images and used as a conditioning for generating images containing those instances. Prior works, such as SlotDiffusion \cite{wu2023slotdiffusion} and Stable-LSD \cite{jiang2023object}, use Stable Diffusion (SD) \cite{rombach2022high} as a decoder to reconstruct images from slot vectors. These models condition the U-Net in SD on slots obtained from Slot Attention, enabling slot-conditional generation.
However, these approaches are fundamentally limited: 1) To control the instances in the generated image, the user needs to manually reconstruct the masks corresponding to all the slots and find target slots that correspond to instances of interest. 2) They fix the diffusion model to a fixed number of slots, as in Slot Attention, limiting flexibility. Users cannot condition on a subset of objects while leaving the image layout flexible, and text inputs are unsupported.\looseness=-1

\method addresses the above limitations: 1) Language-based control is inherent to \method. Hence, to control the instances present in the generated image, a user needs only to query \method to extract the corresponding instance representations from a given image as shown in \fig[(a)]{fig:downstream_tasks}. 2) We maintain the text interface of SD and only add the slots to control the visual identity of specific instances. This is similar to ~\cite{paint_by_example} where a user can specify an image containing an object for instance controlled generation. Instead of specifying images, we use slots obtained from \method for this task. For example, 
if we want to generate a specific bus as in \fig[(a)]{fig:downstream_tasks}, we use \method to extract its representation from an image, allowing us to prompt the diffusion model with ``A photo of a bus. $S_{bus}$'', where $S_{bus}$ is obtained from \method.\looseness=-1

\textbf{\method-SD} We present our image generation pipeline in \fig[(a)]{fig:downstream_tasks}, where \method is kept frozen and SD \cite{rombach2022high} serves as the generative model. We finetune the U-Net in SD, using the COCO dataset, while keeping other components fixed. Captions are used as conditioning inputs, and category names are extracted to obtain corresponding slots via \method. These slots are projected into CLIP embeddings and appended to the caption, as shown in \fig[(a)]{fig:downstream_tasks}. Unlike Stable LSD, where the slot attention module is trained alongside the diffusion model, we treat generation as a downstream task for \method.

\begin{figure*}[]
    \centering
    \small
    \includegraphics[width=0.88\linewidth]{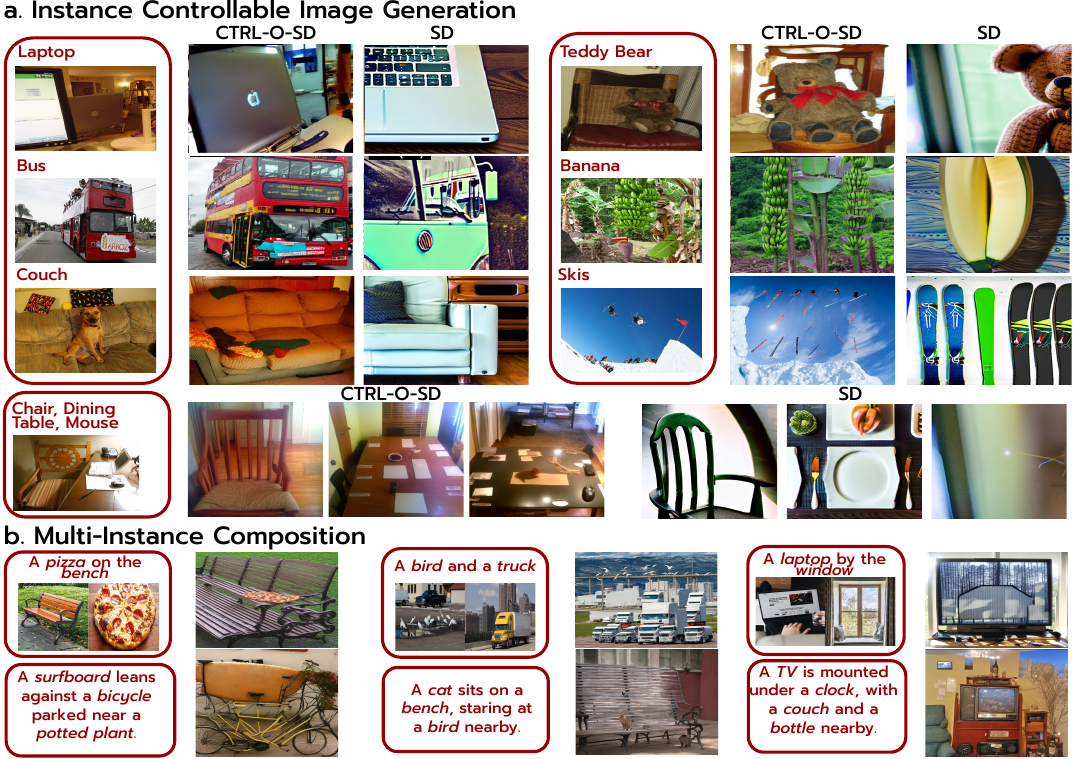}
    \caption{\textbf{a. Instance Controllable Image Generation.} Comparison between \method-SD and the baseline Stable Diffusion (SD). For a given query image (marked \textcolor{stanfordred}{query} ), we extract a slot representation of a specific instance $I_q$ (e.g., laptop, bus, banana). In \method-SD, the input is ``A photo of $I_q$. $S_{I_q}$'' to guide instance generation, while for SD, only ``A photo of $I_q$'' is used. Our approach produces images that more closely match the visual identity of the conditioned instance. \textbf{b. Multi-Instance Composition}. We extract instances from multiple images (e.g., ``bench'' and ``pizza'') and compose them into a single image, as seen with ``the pizza on the bench''.\looseness=-1}
    \label{fig:slot_diffusion}
\end{figure*}

\begin{table}[h]
    \centering
    \begin{minipage}{0.40\linewidth}
        \centering
        \small
        \begin{tabular}{lc}
            \toprule
            \textbf{Method} & \textbf{FID $(\downarrow)$ } \\
            \midrule
            LSD~\citep{jiang2023object}   & 26.20\\
            \method  & \highlight{25.20} \\
            \bottomrule
        \end{tabular}
        \caption{\textbf{Image generation quality}. \method achieves a lower FID.}
        \label{tab:fid_scores}
    \end{minipage}
    \hfill
    \begin{minipage}{0.53\linewidth}
        \centering
        \small
        \begin{tabular}{lc}
            \toprule
            \textbf{Method} & \bf CLIP-I $(\uparrow)$ \\
            \midrule
            SD~\cite{rombach2022high}  & 0.71  \\
            \method & \highlight{0.78} \\
            \bottomrule
        \end{tabular}
        \caption{\textbf{Instance controllable generation}. \method generates instances closer to the query image.}
        \label{tab:clip_score}
    \end{minipage}
    \vspace{-0.5em}
\end{table}



We compare \method-SD to Stable LSD, an existing object-centric method, using Fréchet Inception Distance (FID) to measure generative quality. Since Stable LSD lacks controllability, we focus only on image generation quality. As shown in \Tab{tab:fid_scores}, \method-SD achieves a lower FID, indicating superior image generation.

Next, we assess controllability on the COCO validation set, using SD for comparison. For SD, captions are used as conditioning; for \method-SD, captions and slots extracted from ground truth images are conditioned on object categories. We use the CLIP-I Score, which measures cosine similarity between the CLIP vision embedding of generated and ground truth images. \Tab{tab:clip_score} shows that \method-SD outperforms SD, demonstrating better instance controllability.\looseness=-1


We also visualize the generated images from \method-SD and SD in \Fig{fig:slot_diffusion}. We can see that the images generated using \method-SD contain instances that are closer to the query image. Additionally, \method-SD can compose slots corresponding to two different categories ``A bench'' and ``A pizza'' from two different images, to produce an image specified by ``A pizza on a bench. $S_{Bench}$ $S_{Pizza}$''. We also highlight failure modes in \app{app:ctrlo_sd_failures}, where \method struggles with object binding and the diffusion model faces issues like object deformation and repetition.


\subsection{Visual Questions Answering}
\label{sec:vqa}
\begin{table}[h]
    \centering
    \small
    \begin{tabular}{lcc}
    \toprule
        \bf Model & \bf No-Coupling & \bf Coupling \\
        \hline
        DINOv2 (22M) & 58.26 & 58.12 \\
        CLIP (191M) & 58.43 & 58.64 \\
        DINOSAUR (37M) & 58.32 & 57.66 \\
        \method (39M) & \highlight{59.18} & \highlight{60.25} \\
        \hline
    \end{tabular}
    \caption{\textbf{Performance on VQAv2.} \method achieves the highest accuracy in both settings.\looseness=-1}
    \label{tab:vqa_res}
    \vspace{-0.5em}
\end{table}
We consider VQA task on the VQAv2 dataset~\cite{goyal2017making} as a classification problem, using accuracy as the metric. Language-guided object-centric representations from \method can potentially provide a much stronger coupling between the vision and language inputs. To achieve this, we propose to directly insert the slots into the question before feeding it into the language model (see \fig[(b)]{fig:downstream_tasks}). Given a question, we first use spaCy to extract noun chunks (e.g., Cat, Couch) for the image, which are then used to condition the slots in \method to extract the corresponding slots. These slots are then inserted into the question at the appropriate positions as shown in \fig[(b)]{fig:downstream_tasks}. For example, the question ``What is the color of the cat sitting on the couch?'' becomes ``What is the color of the cat $S_{cat}$ sitting on the couch $S_{couch}$?''. We use a learnable linear projection to map the slots to the same dimension as the T5 embeddings. We feed this question, with the inserted slots, into the language model. Therefore, the language and vision inputs are strongly coupled from the input stage, allowing for more interaction between visual and language components. We refer to this technique as \textit{coupling}. 


\textbf{Setup.} Our full pipeline is shown in \fig[(b)]{fig:downstream_tasks}, inspired by \cite{mamaghan2024exploringVQA}. We use T5 as the language embedding model, and the vision model is \method. For the baselines, we use CLIP \cite{radford2021learning}, DINOv2 \cite{oquab2023dinov2}, and \DINOSAUR \cite{seitzer2023bridging} features as representations. The output network is a transformer with 2 layers and 64 heads. For all methods, we feed visual representations (patches for CLIP and DINOv2 and slots for \method and \DINOSAUR) and the language embeddings from T5 into the output network. The output network is trained from scratch, while T5 undergoes fine-tuning.

We introduce two variants of VQA training with \method: 1) \method that directly feeds the slots into the Transformer output network without embedding them in the language, similar to baseline methods; 2) \method~(with coupling) that inserts the corresponding slot representations into the appropriate place in the question as shown in \fig[(b)]{fig:downstream_tasks}. To ensure a fair comparison, we extend coupling experiments to all baselines. Since these models lack explicit object binding, we insert their aggregated features (CLS token for DINOv2/CLIP and slot mean for DINOSAUR) into the same positions as \method’s control slots. 
We report VQA classification accuracy (see details in~\app{app:vqa_details}).
\looseness=-1

\textbf{Results~(\Tab{tab:vqa_res})}
Both \method variants outperform all baselines, demonstrating the effectiveness of its object-centric representations. Notably, coupling primarily benefits CTRL-O, as its representations are explicitly aligned with language. In contrast, baselines insert full image representations that do not explicitly correspond to the preceding text, limiting coupling's usefulness. We note that while our method improves performance, it remains below the latest state-of-the-art models (>80\%) \cite{liu2023visual}, which leverage large language models (LLMs) and web-scale multimodal data.
\looseness=-2

\section{Conclusion}
We introduced \method, a controllable object-centric model that can be queried to extract representations of specific objects in a scene. We experimentally showed that representations of specific objects can be extracted in complex real-world scenes based on a range of user queries such as object category names or referring expressions. This capability expands the applicability of object-centric models to various real-world applications, 
such as instance-controllable image generation and visual question answering. Therefore, our work takes a step towards improving the applicability of object-centric representations to complex real-world downstream tasks. 
Future work could further enhance object grounding, for example, by considering multi-modal~\citep{jose2024dinov2}, 3D- and motion-aware dense features~\citep{Salehi_2023_ICCV, el2024probing}, making object-centric representations more useful for diverse downstream tasks. We hope that learning controllable object-centric representations becomes the standard in OCL
and leads to broader adoption of object-centric models to various domains and downstream tasks.
\looseness=-1

\section*{Acknowledgments}
The authors would like to thank Kanishk Jain, Ankur Sikarwar, and Le Zhang for reviewing and providing feedback on an earlier version of the paper and Amir M. Karimi and Samuele Papa for providing the code for the VQA experiments in Section \ref{sec:vqa}. This research was enabled in part by compute resources provided by Mila (mila.quebec). AD would like to thank Nanda Harishankar Krishna, Moksh Jain, and Rohan Banerjee for help with Fig. \ref{fig:downstream_tasks}. Additionally, AD would like to thank Anirudh Goyal and Mike Mozer for helpful discussions regarding the research direction explored in this paper.  AD also acknowledges the support of a scholarship from UNIQUE (https://www.unique.quebec). AZ is funded by the European Union (ERC, EVA, 950086). During this project, Aishwarya Agrawal was supported by the Canada CIFAR AI Chair award.

\section*{Contributions}

AD and RA initially started a collaboration on object-centric models for vision-language tasks. AD came up with the idea of controllable slots conditioned on language. AZ proposed the contrastive objective for the model. AD and AZ developed the code for the model and conducted experiments on object grounding. AD ran all the baselines in Section \ref{sec:gounding}. RA developed the code and ran experiments for instance-controllable image generation. AZ and RA developed the initial code for the VQA experiment, which AD and RA further extended for the final results. AZ, RA, and AD ran the baselines for Section \ref{sec:vqa}. AA and AZ provided supervision and guidance throughout the project. MS and EG took part in discussions. AD, RA, AZ, and AA wrote the paper with help from MS.

{
    \small
    \bibliographystyle{ieeenat_fullname}
    \bibliography{main}
}

\clearpage
\setcounter{page}{1}
\maketitlesupplementary
\appendix

{
    \begin{center}
    \end{center}
    \FloatBarrier
}

\section{DINOSAUR Implementation Details} \label{sec:dinosaur}

\begin{figure*}[h]
\centering
\includegraphics[width=0.7\textwidth]{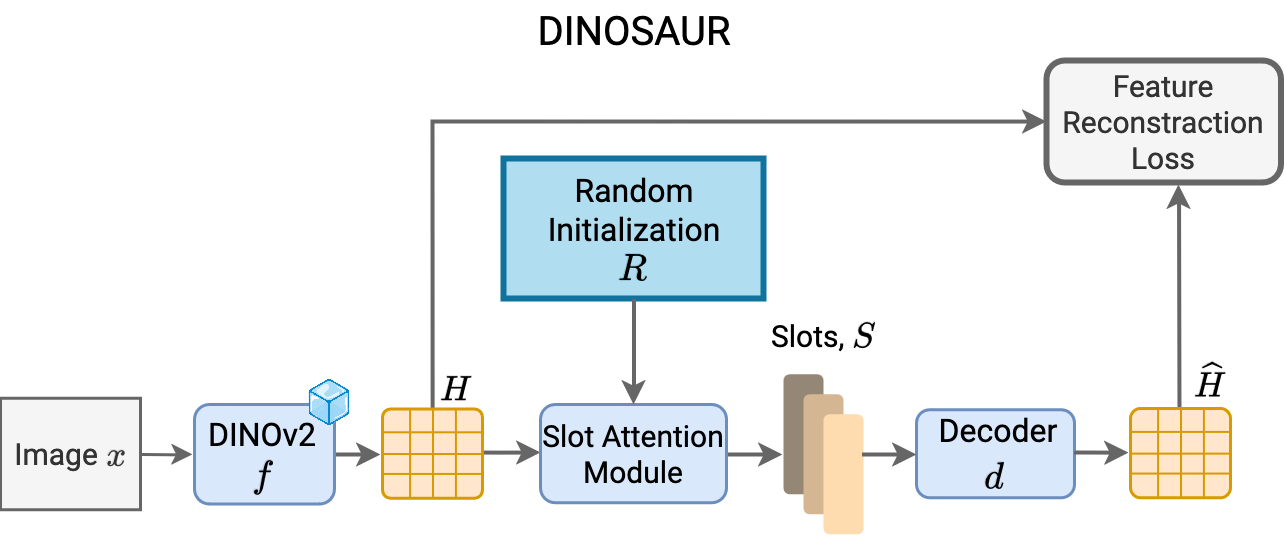} \\
\caption{Overview of \DINOSAUR architecture. The image is processed into a set of
patch features $H$ by a frozen DINO ViT model. The Slot Attention module groups the encoded features
into a set of slots initialized by random queries sampled from the same Gaussian distribution with learnable parameters. By contrast, \method is initialized by the combination of control queries for conditioned slots and random queries for unconditioned slots.   \DINOSAUR  is trained by reconstructing the DINO features from the slots using MLP decoder~\cite{seitzer2023bridging}. }
\label{fig:dinosaur}
\end{figure*}

 \DINOSAUR uses a DINO~\cite{Caron2021DINO} encoder to process the image into features. It relies on a feature reconstruction loss to supervise the object discovery process. Throughout the training, the DINO encoder is kept frozen. We adopt a similar approach, however we use a DINOv2~\cite{oquab2023dinov2} encoder instead of a DINO encoder. \Fig{fig:dinosaur} illustrates the \DINOSAUR architecture with a DINOv2 backbone. Additionally, we have added a learnable mapping network $g$, which is a 3-layer Transformer after the frozen DINOv2 encoder. SA module is applied on top of the mapping output as shown in \Fig[(a)]{fig:framework}.\looseness=-1

\section{\method Implementation Details} \label{app:sec:implement_details}

\begin{algorithm}[H]
\caption{Slot Attention with Language Conditioning}
\label{alg:slot_attention_language}
\textbf{Input:} $\text{inputs} \in \mathbb{R}^{N \times D_\text{inputs}}, \text{slots} \in \mathbb{R}^{K \times D_\text{slot}}, \text{language queries } \ell \in \mathbb{R}^{M \times D_\text{lang}}$ \\
\textbf{Layer params:} $k, q, v$: linear projections for attention; $p_\ell$: projection for language query; GRU; MLP; LayerNorm (x3)
\begin{algorithmic}[1]
\State $\text{inputs} \gets \text{LayerNorm}(\text{inputs})$
\State $\ell_\text{proj} \gets p_\ell(\ell)$ \Comment{Project $M$ language queries to slot dimension}
\State $\text{\{slots\}}_{i=1}^{M} \gets \ell_\text{proj}$ \Comment{Condition first $M$ slots on language query}
\For{$t = 0 \dots T-1$}
    \State $\text{slots}_\text{prev} \gets \text{slots}$
    \State $\text{slots} \gets \text{LayerNorm}(\text{slots})$
    \State $\text{attn} \gets \text{Softmax}(\frac{1}{\sqrt{D}} k(\text{inputs}) \cdot q(\text{slots})^\top$   $\text{axis}=\text{slots} )$
    \State $\text{updates} \gets \text{WeightedMean}(\text{weights}=\text{attn} + \epsilon, \text{values}=v(\text{inputs}))$
    \State $\text{slots} \gets \text{GRU}(\text{state}=\text{slots}_\text{prev}, \text{inputs}=\text{updates})$
    \State $\text{slots} \gets \text{slots} + \text{MLP}(\text{LayerNorm}(\text{slots}))$ 
\EndFor
\State \textbf{return} $\text{slots}$
\end{algorithmic}
\end{algorithm}

We present the modified Slot Attention with query-based initialization in \Alg{alg:slot_attention_language}.  

\paragraph{Control Contrastive Loss} For conditioning, we mainly use language queries. However, we assume that each image in our dataset consists of multiple object annotations, each containing a center of mass annotation and a category or referring expression annotation. Therefore, we have two separate contrastive losses -  one each for the language information and the point information, as shown in \Fig[(b)]{fig:framework}.

\paragraph{Conditioning} We run Slot Attention for a fixed number of slots $K$. However, in general, we may not have $K$ queries per image. In such cases, we initialize a subset of the slots with the given queries, and the rest are free to bind to any of the other objects in the scene~(see line 3 of \Alg{alg:slot_attention_language}). When computing the contrastive loss, we only consider slots conditioned on some query.

\begin{figure*}
    \includegraphics[width=\linewidth]{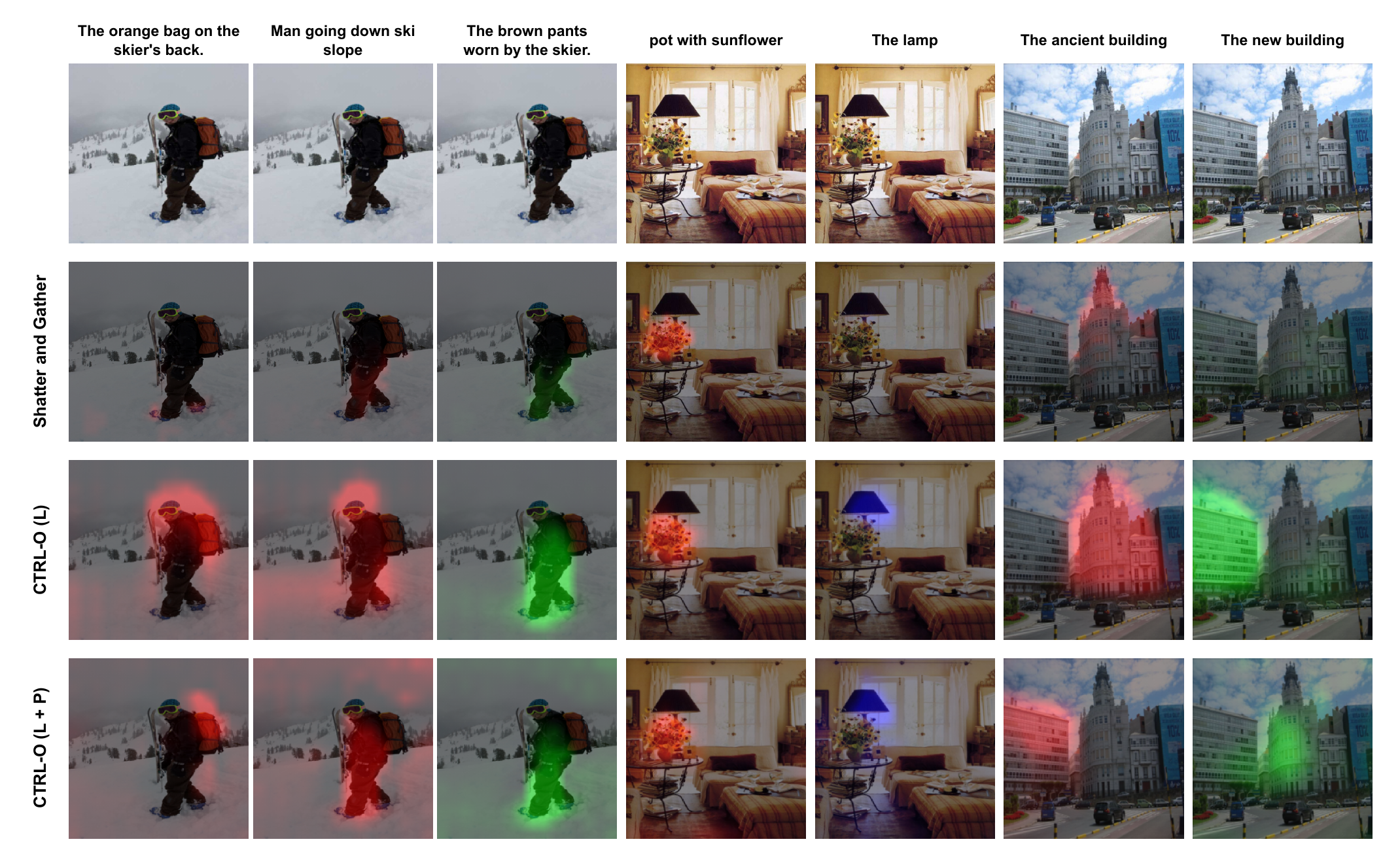}
    \includegraphics[width=\linewidth]{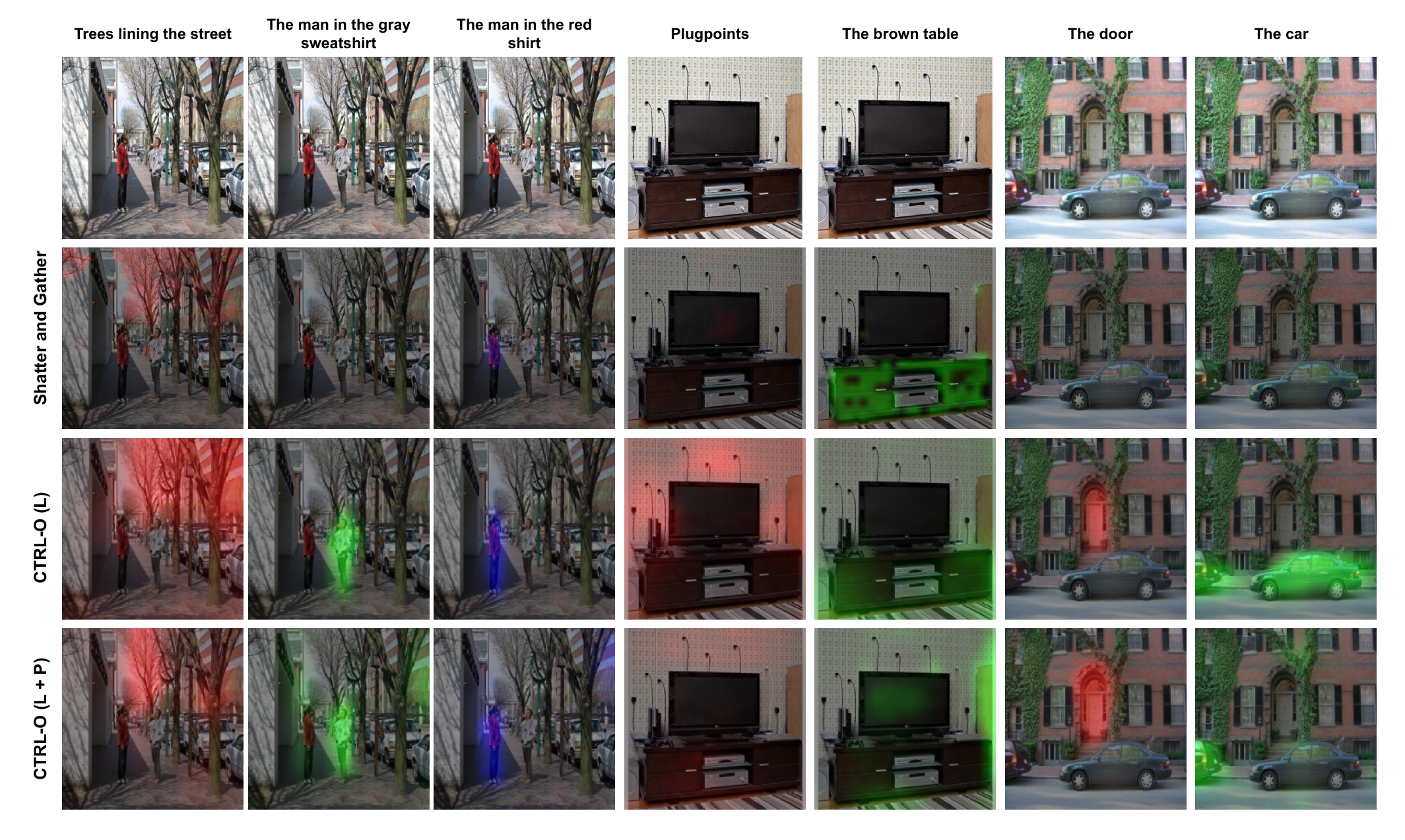}
    \caption{\textbf{Visualization Comparison} In this figure we visualize and compare the masks obtained using \method ($\mathcal{L} + \mathcal{P}$), \method ($\mathcal{L}$), and SaG when queried with free-form language queries. We can see that both the variants based on \method are significantly better at binding to the correct region descriptions as compared to SaG. This difference can be attributed \method using the language guidance to directly influence the slot extraction process while SaG considers the langauge to slot binding as a post-processing step after the slots have been extracted. }
    \label{fig:viscomparison}
\end{figure*}

\section{Training \method with Language Queries}
\label{app:ctrl_o_language_only}
Needing center of mass annotations for the contrastive loss can be a limitation as these annotations may not be available in many datasets. Further, the main baseline that we consider for the referring expression segmentation task (\Sec{sec:gounding}) - Shatter-and-Gather \cite{kim2023shatter} - does not require center of mass annotations. Therefore, for an apples-to-apples comparison, we implement a variant of \method which does not require center of mass annotations.

\begin{figure*}[h]
\centering
\includegraphics[width=0.8\textwidth]{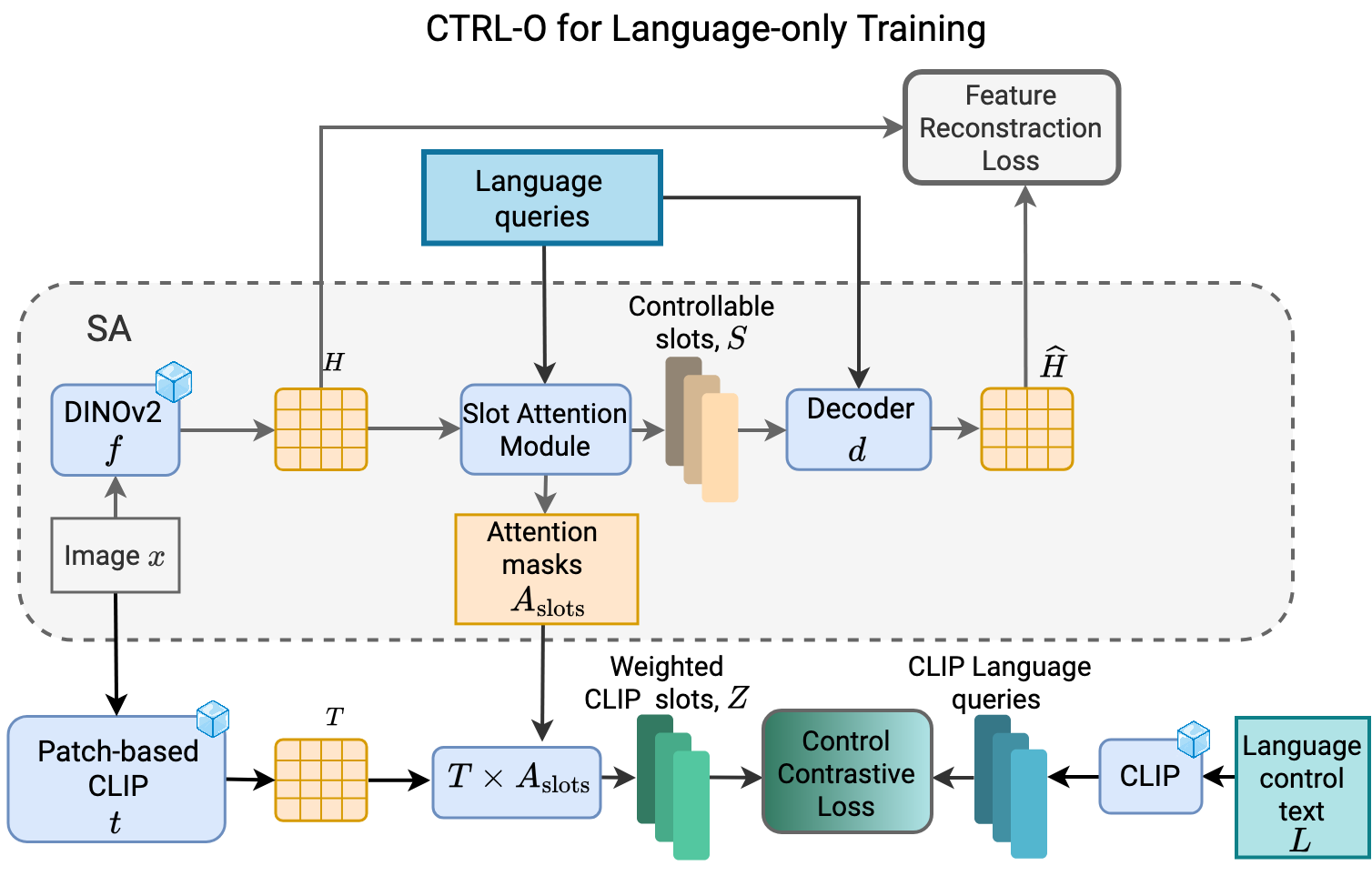} \\
\caption{\textbf{Language-Only \method} Overview of language-only training. In this setting, we use the frozen CLIP model to compute both weighted CLIP slots and CLIP Language queries that we use in control contrastive loss. We average features from CLIP using attention weights from the Slot Attention module. }
\label{fig:langonly}
\end{figure*}

A visual depiction of this approach is presented in \Fig{fig:langonly}. First, we remove additional center-of-mass information and leave only language queries in the contrastive loss. We find that simply removing the center-of-mass information leads to collapse of representations as the contrastive loss can be trivially satisfied by  directly using the language embeddings on which the slots are conditioned on - we term this as \textit{leakage}. To prevent leakage we propose to use CLIP~\citep{radford2021learning} image features and language embeddings in the control contrastive loss. In particular, instead of taking the weighted average of DINO features (\Fig[(a)]{fig:framework}), we take the weighted average of patch-based CLIP features~\citep{fan2023unsupervised}. The slot conditioning still uses language embeddings from LLaMa-3-8B \cite{behnamghader2024llm2vec}, however, CLIP language embeddings are used as targets in the contrastive loss. This way, \method learns to bind to the correct regions in the image specified by language queries without center-of-mass annotations. 

\section{Choice of Decoder for \method}
In this subsection, we additionally study the compatibility of \method with different previously proposed decoders. In particular, we investigate the compatibility and scalability of our method with two different decoder architectures (MLP and Transformer). In Table~\ref{tab:segmentation_comparison}, we compare our method with other OCL methods, showing that while our method strongly outperforms other methods in FG-ARI, its mask quality is lower than methods with stronger (pretrained) diffusion and Transformer decoders that have less inductive bias towards scene decomposition. Thus, it is important to investigate how our method performs with different decoders and whether we can scale MLP decoders for better mask quality.
Object discovery with the Transformer decoder was shown to be sensitive to hyperparameters and can entirely fail (see App. D.4 and D.5 of DINOSAUR paper~\citep{seitzer2023bridging}). Subsequently, we also find that \method with Transformer decoder achieves $10.2$ mBO. Through thorough investigation, we conclude that \emph{Transformer decoder is not compatible with contrastive loss, which is needed for language controllability in CTRL-O but not in the baselines}. Thus, to improve masks quality we propose to scale the MLP decoder itself; scaling MLP dim to 4096 led to improved $28.0$ mBO and $47.9$ FG-ARI.

\section{Referring Expression Visualization}
In \Fig{fig:viscomparison}, we compare the visualizations obtained from \method ($\mathcal{L} + \mathcal{P}$ setting), \method ($\mathcal{L}$ setting), and Shatter-and-Gather (SaG). Note that the queries listed on the top of each column are free-form queries created by a user and may not be similar to those typically found in the visual genome dataset. One potential issue with Shatter-and-Gather is that the language queries do not influence the slot extraction process - Slot attention first extracts a fixed number of slots, after which the query binds to the most relevant slot post-hoc. This can be limiting, as in some cases, the region referred to by the query may not be extracted into a single slot. In such cases, the language query may not bind to any slot. In \Fig{fig:viscomparison}, we find that this is exactly what happens in several cases for Shatter-and-Gather. For example, in the first column, for the query ``The orange bag on the skier's bag'', SaG binds to the skier's shoes. In the 5th column, SaG fails to bind to any region for the query ``the lamp''. In contrast, both variants of \method frequently bind to the correct regions specified by the queries. Secondly, in \method the language queries directly influence slot extraction which allows it to explicitly extract the referred regions from the image and bind to them.

A particularly interesting case is the last row for \method ($\mathcal{L}$), where it learns to bind correctly even though queries are less specific and more subjective - ``the ancient building'' and ``the new building''. This emphasizes the generalizability of \method to complex language queries. 

\section{Object Discovery and Binding Metrics}
\label{app:sec:metrics}

\paragraph{FG-ARI} 
The \textit{adjusted rand index} (ARI) measures the similarity between two clusterings \citep{hubert1985comparing}. We use the instance/object masks as the targets. 
We only compute this metric for pixels in the foreground (hence, FG-ARI). 
Unlabeled pixels are treated as background. 

\paragraph{mBO} 
To compute the mBO~\citep{Arbelaez2014MCG}, each ground truth mask (excluding the background mask) is assigned to the predicted mask with the largest overlap in terms of IoU.The mBO is computed as the average IoU of these mask pairs. 

\paragraph{Binding Hits} This metric measures controllable grounding. For binding hits, consider that a slot $s_i$ is conditioned on a query $L_i$ identifying an object $o_i$ with ground-truth mask $m_i$. The broadcast decoder of slot attention outputs a mask per slot. If the overlap between the predicted mask for slot $s_i$, denoted as $\hat{m}_i$, and the ground truth mask $m_i$ is the highest among all pairs of predicted and ground truth masks, it is considered as a hit (1) else it is considered as a miss (0). Binding Hits metric is measured as the average number of hits across the dataset.

\section{Additional Details of VQA Experiments}
\label{app:vqa_details}
\paragraph{Evaluation Metric.} We evaluate VQA models using classification accuracy across 3000 classes, using the top-frequent answers, which covers more than 90\% of the question in the dataset.

\paragraph{Discussion on Coupling in \method VQA model} The standard approach for solving VQA tasks with pretrained vision and language backbones is to feed the output representations of the vision model and the language model into a single neural network - usually a Transformer~\cite{vaswani2017attention} - which then outputs a distribution over the answer categories \cite{mamaghan2024exploringVQA, manas2022mapl, eichenberg2021magma}. To solve VQA, it is crucial to have strong interaction between the visual and language inputs. However, in pre-existing approaches, this interaction only happens in the output network (the Transformer that processes the language and vision outputs), which can be limiting.  

To address this, we introduce an approach called \textit{coupling}. Coupling, with the help of \method, directly inserts the visual representations into the language query, thus enabling strong vision and language interaction from the input stage. The proposed approach is presented in \Fig[(b)]{fig:downstream_tasks}. %


\section{CTRL-O SD} \label{app:ctrlo-sd}

\subsection{Fine-Tuning Details}
In \method-SD, we finetune a pretrained Stable Diffusion model initialized from the \texttt{stabilityai/stable-diffusion-2-1} checkpoint. As illustrated in \Fig[(a)]{fig:downstream_tasks}, \method extracts slots from a given image based on user-provided queries. These extracted slots are then incorporated into the caption, which is fed into Stable Diffusion. Notably, \method remains frozen during the fine-tuning process, distinguishing our approach from prior works like Slot Diffusion \cite{wu2023slotdiffusion} and Stable LSD \cite{jiang2023object}, where the object-centric model and the diffusion model are trained jointly.

\paragraph{Implementation Details} We train the model on the COCO 2017 training set. For each image, we first extract COCO categories from its associated caption and use these categories to query \method, to generate the corresponding slots. These slots are subsequently appended to the caption, as shown in \Fig[(a)]{fig:downstream_tasks}. The resulting caption is then passed through the CLIP language encoder to condition Stable Diffusion. To integrate \method outputs into the CLIP language embedding space, we introduce a learnable linear layer that maps the extracted slots to the CLIP embedding space. During training, the only components updated are the U-Net parameters of Stable Diffusion. We use random flips as a data augmentation strategy. Training is performed for 300 epochs with a learning rate of \(2 \times 10^{-5}\), utilizing gradient accumulation with 2 steps. Additionally, we reproduce Stable LSD using the author-provided code and hyperparameters on the COCO dataset. The input resolution to the vision encoder for \method is \(224 \times 224\), while Stable LSD uses \(448 \times 448\).

\subsection{Image Generation Metrics}
\label{app:diffusion_metric}

\paragraph{Fréchet Inception Distance (FID) score} We calculate the Fréchet Inception Distance (FID) score~\cite{heusel2017gans} to assess the quality of generated images in comparison to real images. The FID score computes the Fréchet distance between feature distributions of generated and real images, extracted via an Inception v3 model. Lower FID scores indicate a closer match to real images, corresponding to higher image fidelity and diversity. %

\paragraph{CLIP-I Score} We use CLIP-I Score to verify whether the generated images contain the same instances present in the query image. This should be the expected behavior of \method-SD when conditioned on a caption containing slots corresponding to specific instances. We compute this metric on the COCO validation set. We embed the generated image and the query image into the CLIP embedding space using the CLIP ViT Encoder (\texttt{openai/clip-vit-base-patch16}). We then compute the cosine similarity between the two embeddings. This similarity is averaged across all images to compute the final CLIP-I Score.

\subsection{Image Reconstruction Visualization} \label{app:diffusion_reconstruction}

\begin{figure*}
    \centering
    \includegraphics[width=0.87\linewidth]{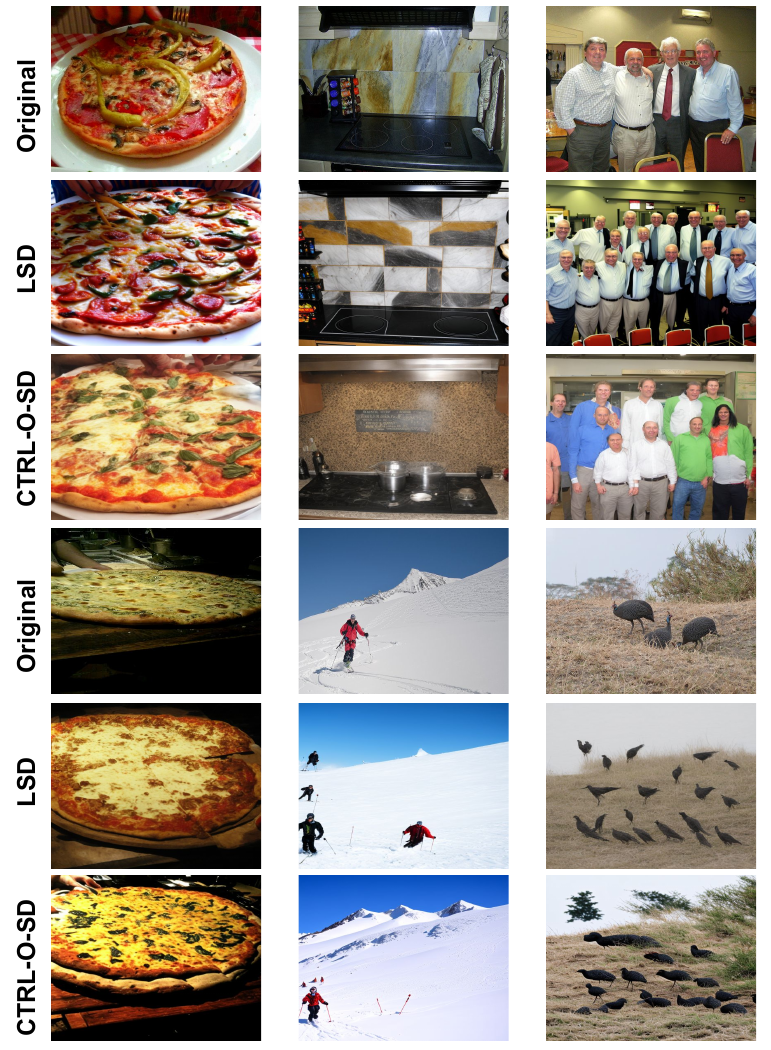}
    \caption{\textbf{Image Reconstruction} Qualitative comparisons of reconstruction outputs for LSD and CTRL-O-SD models. Each column corresponds to a different image. Rows correspond to original inputs, LSD generations, and CTRL-O-SD generations respectively. LSD generates outputs conditioned on full 7-slot representations derived from the original image, while CTRL-O-SD uses captions appended with a subset of slots for conditioning. The results show that both models achieve similar reconstruction quality. }
    \label{fig:reconstruction_output}
\end{figure*}

In this section, we present a qualitative analysis of the  reconstruction capabilities of the LSD and CTRL-O-SD models. The goal is to evaluate how effectively these models retain structural and semantic details. LSD provides the full 7-slot representation derived from the object-centric model to the generative model, providing comprehensive image context for reconstruction. In contrast, CTRL-O-SD provides the caption along with only a subset of slots corresponding to the categories in the caption to the generative model. To obtain these slots, we condition the slots in \method with the categories present in the caption and append the corresponding slots to the caption.   This flexibility in \method-SD enables instance-specific image generation (see ~\cref{fig:slot_diffusion} for examples), which is not feasible with Stable LSD. As illustrated in ~\cref{fig:reconstruction_output}, both models demonstrate comparable reconstruction quality.

\label{app:diffusion_reconstruction}

\subsection{Image Generation Failures}
\label{app:ctrlo_sd_failures}
In this section we highlight some failure cases of \method-SD.

\begin{itemize}
    \item \textbf{Incorrect Focus:} The model occasionally fails to accurately prioritize the main objects in the query, often diverting attention to irrelevant elements. For instance, when prompted to generate an image centered around a cell phone, the model might emphasize a person in the background instead. As we have seen from \Tab{tab:grounding_ablation}, \method does not achieve perfect binding. Hence, this failure can be caused by the slots not binding to the correct regions in the image. 

    \item \textbf{Deformed Outputs:} The model sometimes generates distorted representations of people and animals, with unnatural proportions or malformed features. Such deformities highlight limitations in the model's ability to represent detailed anatomy accurately, indicating a need for refined control over complex shapes and structures. This failure may also be attributed to the failures of the underlying generative model rather than \method.  %

    \item \textbf{Object Duplication:} There are instances where the model replicates objects within a single scene, leading to unrealistic and cluttered outputs.
\end{itemize}

These failure modes suggest areas for further improvement for \method and \method-SD, particularly in object binding for \method and image generation quality for \method-SD.

\begin{figure*}[h]
    \centering
    \includegraphics[width=0.87\linewidth]{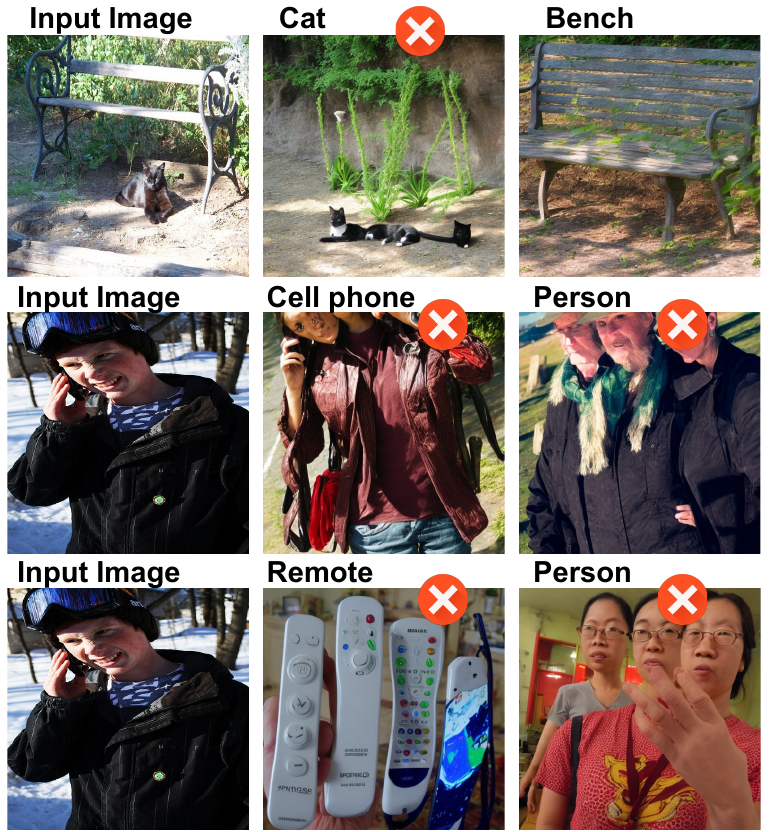}
    \caption{\textbf{Failure Modes of CTRL-O SD.} Examples highlighting some failures in \method-SD such as incorrect focus, deformities in representations of people or animals, and object duplication. Each labeled box illustrates specific instances of these failures.}
    \label{fig:failures}
\end{figure*}

\end{document}